\documentclass[fleqn,10pt]{wlscirep}
\usepackage[utf8]{inputenc}
\usepackage[T1]{fontenc}
\usepackage{amsmath,amssymb}
\usepackage{algorithm}
\usepackage{algpseudocode}
\usepackage{xcolor}
\usepackage{graphicx}
\usepackage{booktabs}
\usepackage{url}
\usepackage{hyperref}

\title{Unified Policy–Value Decomposition for Rapid 
Adaptation}

\author[1,*]{Cristiano Capone}
\author[1,2]{Luca Falorsi}
\author[1]{Andrea Ciardiello}
\author[3]{Luca Manneschi}

\affil[1]{Computational Neuroscience Unit, Istituto Superiore di Sanità, 00161, Rome, Italy}
\affil[2]{Ospedale Santa Lucia, Rome, Italy}
\affil[3]{School of Computer Science, University of Sheffield,Sheffield,S10 2TN, United Kingdom}

\affil[*]{Corresponding author (cristiano.capone@iss.it)}

\begin{abstract}
Rapid adaptation in complex control systems remains a central challenge in reinforcement learning. We introduce a framework in which policy and value functions share a low-dimensional coefficient vector — a goal embedding — that captures task identity and enables immediate adaptation to novel tasks without retraining representations.
During pretraining, we jointly learn structured value bases and compatible policy bases through a bilinear actor–critic decomposition. The critic factorizes as $Q(s,a,g) = \sum_k G_k(g) y_k(s,a)$, where $G_k(g)$ is a goal-conditioned coefficient vector and $y_k(s,a)$ are learned value basis functions. This multiplicative gating — where a context signal scales a set of state-dependent bases — is reminiscent of gain modulation observed in Layer 5 pyramidal neurons, where top-down inputs modulate the gain of sensory-driven responses without altering their tuning~\cite{capone2025adaptive}. Building on Successor Features, we extend the decomposition to the actor, which composes a set of primitive policies weighted by the same coefficients $G_k(g)$. At test time the bases are frozen and $G_k(g)$ is estimated zero-shot via a single forward pass, enabling immediate adaptation to novel tasks without any gradient update.
We train a Soft Actor–Critic agent on the MuJoCo Ant environment under a multi-directional locomotion objective, requiring the agent to walk in eight directions specified as continuous goal vectors. The bilinear structure allows each policy head to specialize to a subset of directions, while the shared coefficient layer generalizes across them, accommodating novel directions by interpolating in goal embedding space.
Our results suggest that shared low-dimensional goal embeddings offer a general mechanism for rapid, structured adaptation in high-dimensional control, and highlight a potentially biologically plausible principle for efficient transfer in complex reinforcement learning systems.
\end{abstract}

\date{}

\begin{document}

\flushbottom
\maketitle
\thispagestyle{empty}

\section{Introduction}

Modern off-policy actor--critic algorithms such as Soft Actor-Critic (SAC) have achieved state-of-the-art performance on continuous control benchmarks by combining stable value estimation, expressive function approximation, and entropy regularization \cite{haarnoja2018soft}. Despite their empirical success, these methods typically rely on \emph{monolithic} neural architectures for both the Q-function and the policy. Such monolithic parametrizations can limit modularity, hinder interpretability, and complicate both theoretical analysis and rapid adaptation when task objectives change.

Motivated by these limitations, a substantial body of work has explored decomposition and modularization in reinforcement learning. Ensemble Q-learning and methods such as bootstrapped DQN or REDQ improve robustness and exploration by aggregating multiple value predictors \cite{osband2016bootstrapped,chen2021redq}. Modular reinforcement learning and mixture-of-experts (MoE) architectures decompose policies or value functions into specialized components coordinated by a learned gating mechanism \cite{andreas2017modular,shazeer2017outrageously}. Successor features (SF) factorize the value function into task-independent features and task-dependent reward weights, enabling efficient transfer across changing reward functions \cite{barreto2017successor}. While these approaches differ in motivation and implementation, they share a common goal: introducing structure into otherwise monolithic function approximators.

In this work, we propose a \emph{co-decomposed bilinear} representation for actor--critic reinforcement learning, in which \emph{both} the critic and the policy are factorized using a shared multiplicative structure. Specifically, the Q-function is represented as
\[
Q(s,a,g) \;=\; \sum_{k=1}^{K} G_k(s,g)\,\phi_k(s,a),
\]
while the deterministic policy is parametrized as
\[
\mu(s,g) \;=\; \sum_{k=1}^{K} G_k(s,g)\,Y_k(s).
\]
Crucially, the same low-dimensional gating vector \(G(s,g)\) modulates both the value components \(\phi_k\) and the policy components \(Y_k\). This \emph{co-decomposition} explicitly couples the actor and critic representations, aligning the directions in which value estimation and policy improvement are expressed.

This design choice departs from standard mixture-of-experts formulations, where the gating function typically serves to \emph{separate} experts or enforce sparsity. Instead, our objective is to \emph{structure and distribute computation} while preserving coherent gradient flow between critic and actor. We show that when the policy and value share the same multiplicative gating, the resulting actor updates are simpler, better aligned with the critic structure, and empirically competitive with significantly more complex monolithic networks.

Beyond learning, the proposed decomposition yields an interpretable and controllable latent space. The learned gating vector \(G\) admits a geometric interpretation, allowing direct modulation of behavior (e.g., direction or speed) by manipulating its magnitude or orientation---including behavioral modes not explicitly encountered during pretraining. Moreover, unlike successor features, our formulation does not assume a known or fixed reward decomposition. Instead, we show that rapid \emph{online adaptation} can be achieved by updating the gating variables directly using a simple value-based rule proportional to the observed reward and the critic components.

The main contributions of this work are summarized as follows:
\begin{itemize}

    \item \textbf{Multiplicative Gating as a Biologically Plausible Bilinear Decomposition:}  
    The proposed architecture relies on multiplicative gating to achieve expressivity, eliminating the need for deep nonlinear activation stacks. This design is biologically plausible, as it resembles gain-modulation mechanisms observed in Layer 5 pyramidal neurons, where contextual inputs multiplicatively modulate ongoing activity. As a result, policies can be represented as multiplicative compositions of linear mappings, enabling simpler networks and facilitating analytical investigation.

    \item \textbf{Policy--Value Co-Decomposition:}  
    We introduce a structured co-decomposition of policy and value functions that shares a latent representation while remaining explicitly separable. This design simplifies optimization compared to standard actor--critic formulations, while achieving comparable performance.

    \item \textbf{Reward-Agnostic Learning Framework:}  
    Unlike Successor Features and related approaches, the proposed method does not assume a predefined or factorized reward structure. Learning proceeds without encoding task-specific reward information into the representation.

    \item \textbf{Behaviorally Interpretable Latent \texorpdfstring{$G$}{G}-Space:}  
    We learn a latent $G$-space that directly modulates policy behavior, providing a compact and interpretable control interface that can be manipulated independently of the policy and value networks.

    \item \textbf{Fast Online Adaptation via \texorpdfstring{$G$}{G}-Space Updates:}  
    We introduce a simple online learning rule operating directly in the $G$-space, enabling rapid behavioral adaptation without retraining the full model and supporting efficient and flexible control.
\end{itemize}

\subsection{Biological Interpretation}

We include a biological parallel as a source of inspiration: cortical circuits achieve rapid context-dependent adaptation, and this motivates our choice of multiplicative modulation in the model.

Figure~\ref{fig1}A should therefore be interpreted as a conceptual analogy, not as a mechanistic claim that our architecture reproduces thalamo-cortical biology. In this analogy, a recurrent processing stage provides a rich spatiotemporal embedding of sensory and internal signals, which can support short-term memory and continuous state representation~\cite{maass2002real, jaeger2001echo, lukosevicius2009reservoir}.

Within this interpretation, two parallel pathways are considered: a context-encoding stream (top), loosely inspired by associative/prefrontal processing, and a subpolicy stream (bottom), loosely inspired by premotor processing~\cite{rigotti2013prefrontal, badre2009prefrontal, wise1985premotor, tanji2001sequential}. Their interaction is mapped to a multiplicative gating motif inspired by dendritic gain modulation in Layer 5 pyramidal neurons~\cite{larkum1999dendritic, hay2011division, urakubo2008nonlinear}.

This biologically inspired viewpoint is used only to motivate architecture design and interpretability: contextual signals modulate action primitives to bind \emph{what action to perform} with \emph{when and under which context}, enabling flexible control without claiming biological equivalence~\cite{larkum2013thalamocortical, sherman2016thalamocortical, rathelot2009motor}.

\section{Results}

\begin{figure*}[t!]
\centering
\includegraphics[width=0.9\linewidth]{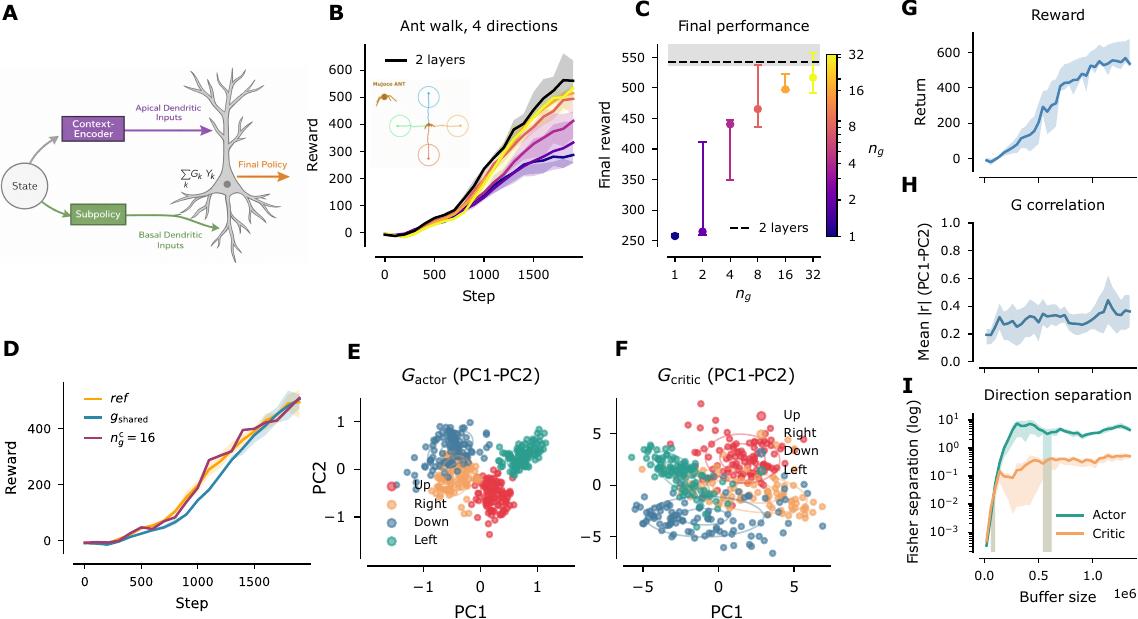}
\caption{\textbf{MLP-based bilinear actor--critic architecture.} 
\textbf{A.} Scheme of our architecture: the actor and critic are decomposed into $K$ parallel basis modules, policy primitives $Y_k(s)$ and value components $\phi_k(s,a)$.  
\textbf{B--C.} Comparison between the learning curves of a traditional architecture (2-layer MLP) and a single-layer MLP with bilinear decomposition. (inset) Scheme of the navigation task: a robot with 8 DOF is asked to move in a specific direction.  
\textbf{D.} Comparison of learning curves between the cases in which the latent space $G_k$ is independent or shared between actor and critic. 
\textbf{E--F.} Direction encoding for actor and critic. 
\textbf{G--I.} Reward, correlation between actor and critic $G$, and direction encoding in the $G$ space, as functions of training steps.}
\label{fig1}
\end{figure*}

We first evaluate the proposed bilinear actor--critic on the directional navigation benchmark and summarize the key outcomes in Fig.~\ref{fig1}. Panel A illustrates the shared-coefficient architecture. Panels B--C report the comparison between a traditional two-layer MLP and a single-layer bilinear model, with the task setup shown in the inset.

The main quantitative result is that bilinear decomposition improves learning efficiency even with a shallower network. As shown in Fig.~\ref{fig1}B--C, the single-layer bilinear model reaches higher reward faster than the standard two-layer MLP baseline, indicating that multiplicative structure can compensate for depth by providing a more task-aligned representation.

We then test the central design choice of this work: sharing the gating coefficients between actor and critic. Fig.~\ref{fig1}D shows that using a common latent vector $G$ yields performance comparable to (or better than) the variant with separate actor/critic gates, while reducing parameterization and optimization complexity.

Finally, Fig.~\ref{fig1}E--I characterizes the learned latent space. Panels E--F show structured direction encoding for actor and critic, while panels G--I report reward, actor--critic $G$ correlation, and the evolution of direction encoding over training. These trends support our claim that shared $G$ forms a coherent control interface that is both behaviorally meaningful and suitable for rapid adaptation. We also observe that actor and critic develop slightly different goal (direction) encodings; when optimized independently, the actor representation can be even more informative for direction decoding. Importantly, constraining the two modules to share the same $G$ does not produce a measurable loss in control performance, while preserving a simpler and more consistent latent interface. We also observe that actor and critic develop slightly different goal (direction) encodings; when optimized independently, the actor representation can be even more informative for direction decoding. Importantly, constraining the two modules to share the same $G$ does not produce a measurable loss in control performance, while preserving a simpler and more consistent latent interface.

We empirically investigated the effect of sharing the $G$ components between the actor and the critic in our proposed bilinear decomposition framework, where
\[
Q(s,a) = \sum_k G_k(s) \, V_k(a,s), \quad \mu(s) = \sum_k G^{(A)}_k(s) \, Y_k(s) .
\]
In the general formulation, the actor and the critic may have different $G$ functions ($G^{(A)} \neq G$), allowing them to be optimized independently. However, our experiments show that setting $G^{(A)} = G$ leads to essentially identical performance compared to learning $G^{(A)}$ separately.  

This finding has two important consequences:
\begin{enumerate}
    \item It is not necessary to optimize $G^{(A)}$ at all; one can directly reuse the critic's $G$ for the actor, thus reducing the number of parameters and simplifying the optimization process.
    \item When facing a new task, adaptation can be achieved by re-estimating only $G$ from the critic, without re-learning $G^{(A)}$ and subsequently re-optimizing the actor. This substantially reduces the cost of transfer to new tasks.
\end{enumerate}

This parameterization of the actor---using the same $G$ factors as the critic within a bilinear decomposition---is, to the best of our knowledge, novel. Our results suggest that it is not only theoretically sound, but also practically beneficial for both computational efficiency and rapid transfer learning.

\begin{figure*}[t!]
\centering
\includegraphics[width=0.9\linewidth]{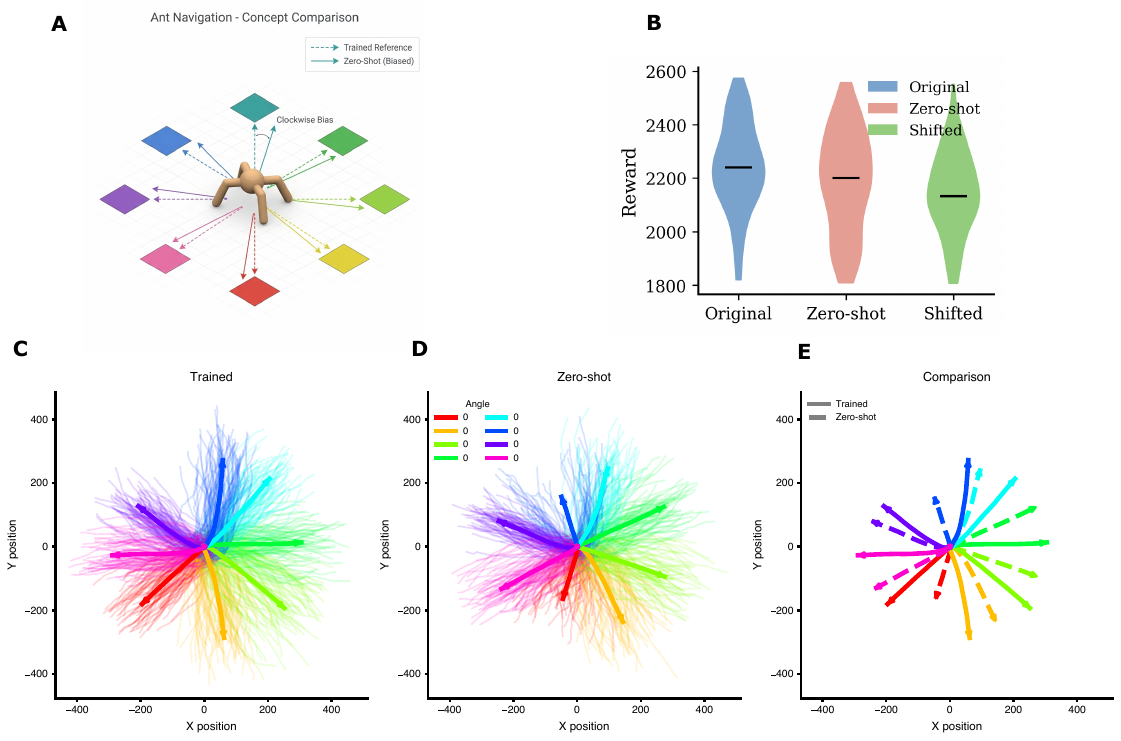}
\caption{\textbf{Zero-shot learning.} \textbf{A.} Task scheme: the MuJoCo Ant agent is pre-trained on target directions and tested on new ones. Pretrained bilinear agent is evaluated on unseen goal directions (or task descriptors) without any parameter update, by conditioning on $g$. \textbf{B.} Performance compared against baselines when switching to novel directions. \textbf{C--D.} Behavior trajectories for training and test directions, respectively, illustrating successful generalization to intermediate angles not explicitly trained. \textbf{E.} Direct comparison between train and test directions (averaged over trials).}
\label{fig2}
\end{figure*}

We next evaluate zero-shot generalization to unseen target directions (Fig.~\ref{fig2}). Panel A summarizes the protocol: the MuJoCo Ant is pretrained only on directional objectives and then tested on new directions without any parameter update, conditioning only on $g$. Panel B shows that the pretrained bilinear agent remains competitive with baseline methods after direction switches. Panels C--D report representative trajectories for training and test directions, showing smooth interpolation toward intermediate headings that were not explicitly seen during training. Finally, panel E compares train and test directional performance averaged over trials, confirming only limited degradation in the zero-shot regime and supporting the view that $G$ acts as a structured control variable rather than a standard contextual input.

\begin{figure*}[t!]
\centering
\includegraphics[width=0.9\linewidth]{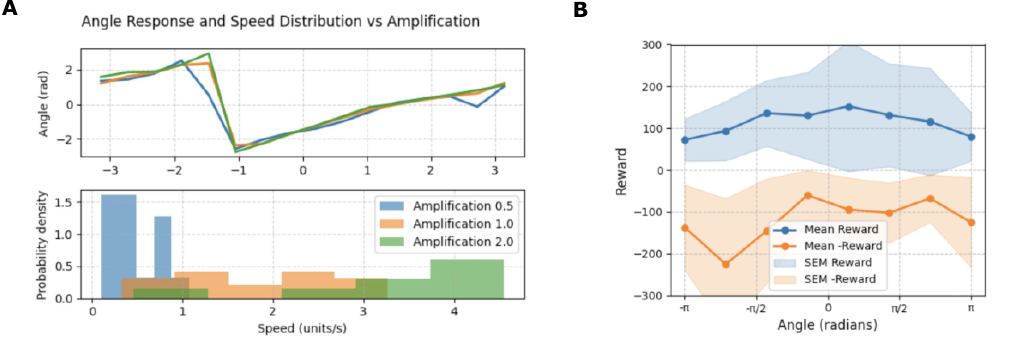}
\caption{\textbf{Bilinear decomposition allows for interpretability and generalization.} \textbf{A.} Manipulating individual gating coordinates $G_k$ produces consistent, semantically meaningful changes in behavior, indicating interpretable control axes. (Top) Average movement direction as a function of latent-space direction for three different amplitudes; (bottom) speed distribution for the same three amplitudes (color-coded). Notably, speed modulation emerges spontaneously, although training objectives were defined only over movement direction. \textbf{B.} Online adaptation of $G_k$, allowing real-time solution of the current task. Reward as a function of the target direction (blue). As a reference, a task with negative reward is shown (orange).}
\label{fig3}
\end{figure*}

Figure~\ref{fig3} highlights the interpretability and online adaptability of the learned gating space. In panel A, controlled manipulation of individual coefficients $G_k$ produces systematic and semantically meaningful behavioral changes: the top plot shows how movement direction varies with latent-space direction at three amplitudes, while the bottom plot reports the corresponding speed distributions. Importantly, modulation in $G$ also produces coherent changes in movement speed, even though speed was never an explicit training target. These results indicate that $G$ acts as a structured control interface rather than an opaque latent code; this behavior was not achievable when the contextual signal was treated as a traditional network input (results not shown).

Panel B demonstrates real-time adaptation through direct updates of $G_k$. As task direction changes, reward remains aligned with the target-direction objective (blue), whereas a negative-reward reference task (orange) provides a clear contrast. Together, these findings support the claim that bilinear co-decomposition enables both interpretable control and fast task-level adaptation without retraining the full network.

\section{Methods}

\subsection{Multiplicative gating enables a bilinear decomposition of the policy}

We consider a deterministic continuous-control policy $\mu(s,g) \in \mathbb{R}^{|\mathcal{A}|}$ conditioned on the environment state $s$ and a low-dimensional goal or context variable $g$. Rather than representing the policy as a single monolithic nonlinear mapping, we decompose it into a set of $K$ basis action generators modulated by a multiplicative gating signal:
\begin{equation}
\mu(s,g) \;=\; \sum_{k=1}^{K} G_k(s,g)\, Y_k(s),
\label{eq:policy_bilinear}
\end{equation}
where $Y_k(s) \in \mathbb{R}^{|\mathcal{A}|}$ are state-dependent basis policies, and $G_k(s,g) \in \mathbb{R}$ are scalar gating coefficients.

This formulation induces a \emph{bilinear} structure: the policy output is linear in the basis functions $Y_k(s)$ for fixed gating, and linear in the gating vector $G(s,g)$ for fixed state features. As a result, expressive policies can be represented through multiplicative interactions without relying on deep stacks of nonlinear activations. In practice, both $Y_k$ and $G_k$ are implemented as shallow neural networks, but their interaction remains explicitly multiplicative.

From a functional perspective, Eq.~\eqref{eq:policy_bilinear} can be interpreted as a low-rank factorization of a general policy mapping. Any policy of the form $\mu(s,g) = f(s,g)$ can be approximated by a finite sum of separable terms $G_k(s,g) Y_k(s)$, with $K$ controlling the rank and expressive capacity. Importantly, this factorization disentangles \emph{what actions are available} (encoded by the basis policies $Y_k$) from \emph{how strongly they are expressed} (controlled by $G_k$).

Multiplicative gating plays a crucial role in this decomposition. Unlike additive mixtures, multiplicative modulation allows the same basis policy to be selectively amplified, suppressed, or combined with others depending on context. This mirrors well-known computational motifs in biological neural circuits, where multiplicative interactions enable contextual control and nonlinear integration while preserving linear readouts.

A further consequence of Eq.~\eqref{eq:policy_bilinear} is interpretability. When the basis policies specialize along distinct behavioral dimensions (e.g., forward motion, turning, speed modulation), the gating coefficients acquire a direct semantic meaning. Manipulating individual components of $G$ produces predictable and smooth changes in behavior, effectively defining a low-dimensional control manifold embedded in the full action space.

Finally, this bilinear policy representation is particularly well-suited for actor--critic learning. As shown in the following sections, sharing the same gating variables between policy and critic induces aligned gradients and enables efficient optimization, while also supporting fast online adaptation by updating only the low-dimensional gating vector.

\subsection{Bilinear Co-Decomposition of Actor and Critic}

We propose a structured \emph{co-decomposition} for both policy and value function in continuous control. The critic $Q(s,a,g)$ is factorized as a weighted sum of component value functions:
\begin{equation}
Q(s,a,g) = \sum_{k=1}^{K} G_k(s,g)\,\phi_k(s,a),
\end{equation}
where $\phi_k$ are basis value functions and $G_k(s,g)$ are gating scalars that modulate their contribution. 

Similarly, the deterministic actor $\mu(s,g)$ is represented as
\begin{equation}
\mu(s,g) = \sum_{k=1}^{K} G_k(s,g)\,Y_k(s),
\end{equation}
where $Y_k(s)$ are basis action generators. Crucially, the same gating vector $G(s,g)$ modulates both actor and critic, creating an explicit \emph{alignment between policy improvement directions and value estimates}. This co-decomposition allows distributed computation across $K$ components while retaining a compact, interpretable latent space $G$.

In our implementation, the gating vector is produced from a small linear layer acting on the goal or directional input of the agent, optionally with injected Gaussian noise for exploration (see Actor and Critic classes). Each component is then scaled by its respective $G_k$ and summed to produce the final action or Q-value.

\subsection{Actor--Critic Training}

We adopt a standard off-policy soft actor--critic (SAC) framework \cite{haarnoja2018soft}, modified to respect the co-decomposed structure. Two critics $Q_1$ and $Q_2$ are maintained, along with their target networks. The actor samples actions according to
\begin{equation}
a \sim \mathcal{N}(\mu(s,g), \sigma(s,g)^2),
\end{equation}
with mean $\mu(s,g)$ given by the bilinear decomposition and state-dependent diagonal covariance $\sigma(s,g)^2$.

Critic updates are performed by minimizing the squared Bellman error:
\begin{equation}
\mathcal{L}_{Q_i} = \mathbb{E}_{(s,a,r,s')} \Big[ \big( Q_i(s,a,g) - y \big)^2 \Big],
\end{equation}
where
\begin{equation}
y = r + \gamma \min_j Q_j(s',a',g'),
\end{equation}
with $a'$ sampled from the current actor and $g'$ the gating vector for $s'$. Actor updates follow the soft policy gradient:
\begin{equation}
\mathcal{L}_\mu = \mathbb{E}_s \big[ \alpha \log \pi(a|s) - \min_j Q_j(s,a,g) \big].
\end{equation}

To enforce co-decomposition, the actor and both critics share the same gating layer, ensuring that the same $G_k$ modulates both policy and value components. Target networks are updated using an exponential moving average.

\paragraph{Shared-Coefficient Policy and Value Decomposition}
We parameterize both the action--value function and the policy using a shared low-dimensional coefficient vector $G \in \mathbb{R}^K$.
Specifically, the action--value function is represented as
\[
Q(s,a) = \sum_{k=1}^K G_k \, \psi_k(s,a),
\]
where $\{\psi_k\}_{k=1}^K$ are fixed successor-feature bases learned during a pretraining phase.
The policy is decomposed analogously as a convex combination of basis policies,
\[
\pi(a|s) = \sum_{k=1}^K G_k \, Y_k(a|s),
\]
where each $Y_k(a|s)$ is a valid stochastic policy.
This construction ensures that the value function corresponds to the policy induced by $G$, preserving policy--value consistency.

\subsection{Zero-Shot Adaptation}

We evaluate zero-shot transfer by freezing all learned parameters and directly conditioning the policy on a new goal descriptor $g^*$. In this setting, the gating network produces coefficients $G(s,g^*)$ with a single forward pass, and the policy is executed without gradient steps or replay-buffer updates:
\[
\pi_{\text{ZS}}(a|s,g^*) = \sum_{k=1}^{K} G_k(s,g^*)\,Y_k(a|s).
\]
This protocol measures how well the pretrained bases and shared coefficient structure generalize to unseen directions/tasks purely through interpolation in the learned latent space. We report zero-shot performance using return and trajectory alignment metrics before enabling any adaptation dynamics.

\subsection{Online Adaptation in G-Space: Method and Derivation}

At adaptation time, we freeze the pretrained bases $\{Y_k,\psi_k\}_{k=1}^K$ and update only the shared coefficient vector
$G \in \mathbb{R}^K$. Using $w \equiv G$, we write
\[
Q(s,a \mid g) = \psi(s,a)^\top w,
\qquad
\psi(s,a) = [\psi_1(s,a),\dots,\psi_K(s,a)]^\top.
\]
Given a transition $(s_t,a_t,r_t,s_{t+1},a_{t+1})$, the TD error is
\[
\delta_t = r_t + \gamma\,\psi(s_{t+1},a_{t+1})^\top w - \psi(s_t,a_t)^\top w.
\]
Minimizing $\frac{1}{2}\delta_t^2$ with stochastic gradient descent yields the linear TD/SARSA update
\[
w \leftarrow w + \alpha_G\,\delta_t\,\psi(s_t,a_t),
\]
which is then immediately reused by the policy decomposition
\[
\pi(a\mid s,g) = \sum_{k=1}^{K} w_k\,Y_k(a\mid s).
\]
This gives rapid adaptation without retraining actor or critic bases.

The simplified rule discussed in the Results and Discussion,
\[
\Delta G_k \propto r\,\phi_k(s,a),
\]
is recovered as a one-step approximation of the TD update when the bootstrap term is neglected over short horizons and $\phi_k$ is identified with the critic basis response.
In practice, we use the TD form above because it is more stable and value-consistent. Importantly, adaptation is value-based and does not rely on policy-gradient updates in $G$-space.

\subsection{Experimental Setup}

We evaluate our approach on the MuJoCo \texttt{Ant-v4} environment. The agent is trained to move in a given target direction, encoded as a 2D vector appended to the observation. During training, directions are cycled every 100 steps through eight angles (four cardinal and four diagonal), ensuring exposure to multiple behavioral modes. Episodes last 800 steps, and reward is defined as forward progress along the target direction minus a small penalty for orthogonal movement:
\begin{equation}
r = \Delta x \cos\theta + \Delta y \sin\theta - 0.1\,|\Delta x \sin\theta - \Delta y \cos\theta|.
\end{equation}

We employ a replay buffer of size $10^5$ and train both actor and critic using Adam with learning rate $3\times10^{-4}$. Each training iteration samples a batch of transitions and performs standard SAC updates, while online G-space adaptation is applied during rollout to demonstrate rapid directional modulation.

\subsection{Analysis of G Dynamics}

To analyze the learned and adapted gating vectors, we record $G$ activations during episodes and visualize them using PCA. This provides insight into the monosemanticity of each component (i.e., each $G_k$ tends to correspond to a consistent action pattern) and shows that online adaptation can modulate direction and speed independently of the pre-trained policy, highlighting the interpretability and controllability of the co-decomposition.

\section{Discussion}

Our results provide a coherent picture across Figs.~\ref{fig1}--\ref{fig3}. First, the shared bilinear actor--critic decomposition improves learning efficiency and preserves performance while reducing optimization complexity through shared coefficients. Second, zero-shot transfer to unseen directions is effective without parameter updates. Third, direct manipulations of $G$ reveal an interpretable control space that supports fast online adaptation.

A key outcome is the practical controllability of behavior in $G$-space. Varying $G$ does not only steer movement direction; it also modulates movement speed in a smooth and reliable way. Crucially, this speed control emerges spontaneously: training objectives explicitly targeted direction, not speed. This emergent degree of freedom is important because it indicates that $G$ is not merely a standard contextual input appended to the policy, but a structured latent control interface shared by actor and critic.

From a mechanistic perspective, adaptation is achieved by updating only $G$ while keeping basis functions fixed. Using the value-based rule
\[
\Delta G_k \propto r \, \phi_k(s,a),
\]
reward information is propagated directly through the decomposition, yielding rapid behavioral adjustment within a few environment steps. In our experiments, policy-gradient updates in $G$-space were less effective, supporting the idea that value-based updates are better aligned with this representation.

Overall, the proposed decomposition combines sample-efficient learning, zero-shot generalization, interpretable low-dimensional control, and rapid online adaptation in a single framework. These properties motivate future work on learning bases $\phi_k$ with stronger monosemanticity and extending the same principle to hierarchical and multi-goal control settings.


\bibliographystyle{unsrt}
\bibliography{references.bib}

\end{document}